\documentclass{article}

    \usepackage[preprint,nonatbib]{neurips_2020}

\usepackage[numbers,sort]{natbib}
\usepackage[utf8]{inputenc} 
\usepackage[T1]{fontenc}    
\usepackage{url}            
\usepackage{booktabs}       
\usepackage{amsfonts}       
\usepackage{nicefrac}       
\usepackage{microtype}      
\usepackage{multirow}
\usepackage{graphicx}
\graphicspath{ {./figures/} }
\usepackage{makecell}
\usepackage{authblk}
\usepackage[
  pagebackref,
  breaklinks=true,
  letterpaper=true,
  pageanchor=true,
  plainpages=false,
  pdfpagelabels,
  bookmarks=false,
  bookmarksnumbered,
  colorlinks=true,
  linkcolor=red,
  citecolor=blue,
  urlcolor=magenta,
  pdfborder=0 0 0,  
]{hyperref}

\renewcommand*{\Affilfont}{\normalsize\normalfont}

\title{Training EfficientNets at Supercomputer Scale:\newline 83\% ImageNet Top-1 Accuracy in One Hour}

\author[1, 2, *]{Arissa Wongpanich\thanks{*Work done during Arissa's internship at Google. Corresponding authors: \texttt{arissa@berkeley.edu} and \texttt{sameerkm@google.com}}}
\author[1]{Hieu Pham}
\author[2]{James Demmel}
\author[1]{Mingxing Tan}
\author[1]{Quoc Le}
\author[3]{Yang You}
\author[1]{Sameer Kumar}
\renewcommand\footnotemark{}

\makeatletter
\renewcommand\AB@affilsepx{\hspace{0.02\textwidth} \protect\Affilfont}
\makeatother
\affil[1]{Google Research}
\affil[2]{University of California, Berkeley}
\affil[3]{National University of Singapore}

\begin{document}

\maketitle

\begin{abstract}
EfficientNets are a family of state-of-the-art image classification models based on efficiently scaled convolutional neural networks. Currently, EfficientNets can take on the order of days to train; for example, training an EfficientNet-B0 model takes 23 hours on a Cloud TPU v2-8 node \cite{cloudTPU}.  In this paper, we explore techniques to scale up the training of EfficientNets on TPU-v3 Pods with 2048 cores, motivated by speedups that can be achieved when training at such scales. We discuss optimizations required to scale training to a batch size of 65536 on 1024 TPU-v3 cores, such as selecting large batch optimizers and learning rate schedules as well as utilizing distributed evaluation and batch normalization techniques. Additionally, we present timing and performance benchmarks for EfficientNet models trained on the ImageNet dataset in order to analyze the behavior of EfficientNets at scale. With our optimizations, we are able to train EfficientNet on ImageNet to an accuracy of 83\% in 1 hour and 4 minutes.

\end{abstract}

\section{Introduction}\label{intro}

As machine learning models have gotten larger \cite{gpt3}, so has the need for increased computational power. Large clusters of specialized hardware accelerators such as GPUs and TPUs can currently provide computations on the order of petaFLOPS, and have allowed researchers to dramatically accelerate training time. For example, the commonly used ResNet-50 image classification model can be trained on ImageNet \cite{imagenet} in 67 seconds on 2048 TPU cores \cite{imageClassificationAtSupercomputerScale}, a substantial improvement from a typical training time taking on the order of hours. In order to accelerate training of machine learning models with petascale compute, large-scale learning techniques as well as specialized systems optimizations are necessary. 

EfficientNets~\cite{efficientNet}, a family of efficiently scaled convolutional neural nets, have recently emerged as state-of-the-art models for image classification tasks. EfficientNets optimize for accuracy as well as efficiency by reducing model size and floating point operations executed while still maintaining model quality. Training an EfficientNet-B0 model on Cloud TPU v2-8, which provides 8 TPU-v2 cores, currently takes 23 hours~\citep{cloudTPU}. By scaling up EfficientNet training to a full TPU-v3 pod, we can significantly reduce this training time.

Training at such scales requires overcoming both algorithmic and systems-related challenges. One of the main challenges we face when training at scale on TPU-v3 Pods is the degradation of model accuracy with large global batch sizes of 16384 or greater. Additionally, the default TensorFlow APIs for TPU, TPUEstimator~\citep{tpuestimator}, constrains evaluation to be performed on a separate TPU chip, thereby creating a new compute bottleneck from the evaluation loop~\citep{tensorflow,tfEstimator}. To address these challenges, we draw from various large-scale learning techniques, including using an optimizer designed for training with large batch sizes, tuning learning rate schedules, distributed evaluation, and distributed batch normalization. With our optimizations, we are able to scale to 1024 TPU-v3 cores and a batch size of 65536 to reduce EfficientNet training time to one hour while still achieving 83\% accuracy. We discuss our optimizations in Section \ref{methods} and provide analysis and benchmarks of our results in Section \ref{results}.

\section{Related Work}
Training machine learning models with more TPU cores requires increasing the global batch size to avoid under-utilizing the cores. This is because the TPU cores operate over a memory layout of XLA ~\citep{xla}, which pads each tensor’s batch dimension to a multiple of eight \citep{tpu}. When the number of TPU cores increases to the point that each core processes fewer than 8 examples, the cores will have to process the padded examples, thus wasting resources. Therefore, training on an entire TPU-v3 pod which has 2048 TPU cores requires at least a global batch size of 16384.

It has been observed that when training with such large batch sizes there is degradation in model quality compared to models trained with smaller batch sizes due to a ``generalization gap''~\citep{generalization_gap}. Previous works on large-batch training have addressed this issue using an amalgamation of techniques, such as:
\begin{itemize}
\item Adjusting the learning rate scaling and warm-up schedules \citep{goyal}
\item Adjusting the computation of batch-normalization statistics  \citep{akiba,goyal}
\item Using optimizers that are designed for large batch sizes, such as LARS \citep{lars} or SM3 \citep{sm3}
\end{itemize}
Together, these techniques have allowed training ResNet-50 on ImageNet in 2.2 minutes~\citep{imageClassificationAtSupercomputerScale}, BERT in 76 minutes~\citep{lamb}, and more recently ResNet-50 on ImageNet in under 30 seconds~\citep{mlperfResults}, all without any degradation in model quality.

Despite all the impressive training time measures, we observe that in the image domain, these scaling techniques have merely been applied to ResNets. Meanwhile, these techniques have not been applied to EfficientNet despite their state-of-the-art accuracy and efficiency.

\section{Methods}\label{methods}
Scaling EfficientNet training to 1024 TPU-v3 cores introduces many challenges which must be addressed with algorithmic or systemic optimizations. The first challenge we face is maintaining model quality as the global batch size increases. Since the global batch size scales with the number of cores used for training, we must utilize large batch training techniques to maintain accuracy. We also face compute bottlenecks when training across large numbers of TPU chips, which we address using the distributed evaluation and batch normalization techniques presented in \citet{scale_mlperf}. The optimization techniques we explore to scale EfficientNet training on TPU-v3 Pods are described below:
\subsection{Large batch optimizers} 
While the original EfficientNet paper used the RMSProp optimizer, it is known that with larger batch sizes RMSProp causes model quality degradation. We scale training to 1024 TPU-v3 cores via data parallelism, which means that the global batch size must scale up with the number of workers if we keep the per-core batch size fixed. For example, if we fix the per-core batch size at 32, the resulting global batch size on 1024 cores would be 32768. On the other hand, if the global batch size is fixed when scaling up to many cores, the resulting lower per-core batch size leads to inefficiencies and lower throughput. Thus, large global batch sizes are necessary for us to more optimally utilize the memory of each TPU core and increase throughput. Using the Layer-wise Adaptive Rate Scaling (LARS) optimizer proposed in \citet{lars}, we are able to scale up to a batch size of 65536 while attaining similar accuracies as the EfficientNet baseline accuracies reported in \citet{efficientNet}. 
\subsection{Learning rate schedules}
In order to maintain model quality at large batch sizes, we also adopt the learning rate warmup and linear scaling techniques described in \citep{goyal}. Increasing the global batch size while keeping the number of epochs fixed results in fewer iterations to update weights. In order to address this, we apply a linear scaling rule to the learning rate for every 256 samples in the batch. However, larger learning rates can lead to divergence; thus, we also apply a learning rate warmup where training starts with a smaller initial learning rate and gradually increases the learning rate over a tunable number of epochs. In addition, we compared various learning rate schedules such as exponential decay and polynomial decay and found that for the LARS optimizer, a polynomial decay schedule achieves the highest accuracy.
\subsection{Distributed evaluation}
The execution of the evaluation loop is another compute bottleneck on the standard cloud TPU implementation of EfficientNet, since evaluation and training loops are executed on separate TPUs. With traditional TPUEstimator~\citep{tpuestimator}, where evaluation is carried out on a separate TPU, training executes faster than evaluation, causing the end-to-end time to depend heavily on evaluation time. To overcome this, we utilize the distributed training and evaluation loop described in \citet{scale_mlperf}, which distributes training and evaluation steps across all TPUs and allows for scaling to larger numbers of replicas.

\subsection{Distributed batch normalization}
Additionally, we distribute the batch normalization across replicas by grouping subsets of replicas together, using the scheme presented in \citet{imageClassificationAtSupercomputerScale}. This optimization improves the final accuracy achieved with trade-offs on the communication costs between TPUs. The number of replicas that are grouped together is a tunable hyperparameter. The resulting batch normalization batch size, which is the total number of samples in each replica subset, also affects model quality as well as convergence speed. For subsets of replicas larger than 16, we also explore a two-dimensional tiling method of grouping replicas together.

\subsection{Precision}
It has been observed that using the bfloat16 floating point format for training convolutional neural networks can match or even exceed performance of networks trained using traditional single precision formats such as fp32~\citep{2018choi,mixedPrecision,kalamkar}, possibly due to a regularizing effect from the lower precision. We implement mixed-precision training to take advantage of the performance benefits of bfloat16 while still maintaining model quality. In our experiments, bfloat16 is used for convolutional operations, while all other operations utilize fp32. Using the bfloat16 format for convolutions improves hardware efficiency without degradation of model quality. 
\begin{figure}[htb!]
  \centering
  \includegraphics[scale=0.5]{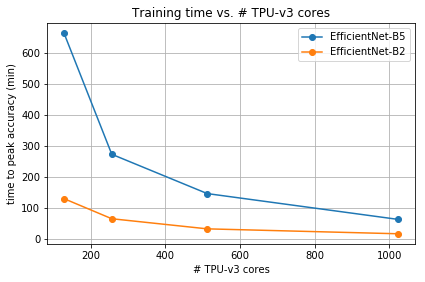}
  \caption{\label{fig:time_to_peak_acc}EfficientNet-B2 and B5 training time to peak accuracy for various TPU slice sizes. Training time starts immediately after initialization of the distributed training and evaluation loop and ends when the model reaches peak accuracy.}
\end{figure}
\section{Results}\label{results}

In this section, we provide results from combining the techniques described above to train a variety of EfficientNet models on the ImageNet dataset at different TPU Pod slice sizes. We train for 350 epochs to provide a fair comparison between the original EfficientNet baseline and our methods. We benchmark training times and accuracy by taking an average of three runs for each set of hyperparameters and configurations. The training time is measured by computing the time immediately after initialization of the distributed training and evaluation loop to the time when peak top-1 evaluation accuracy is achieved. As shown in
 Figure~\ref{fig:time_to_peak_acc}, we are able to observe a training time of 18 minutes to 79.7\% accuracy for EfficientNet-B2 on 1024 TPU-v3 cores with a global batch size of 32768, representing a significant speedup. By scaling up the global batch size to 65536 on 1024 TPU-v3 cores, we are able to reach an accuracy of 83.0\% in 1 hour and 4 minutes on EfficientNet-B5. The full benchmark of accuracies and respective batch sizes can be found in Table~\ref{tab:peak_acc}, demonstrating that with our methods we are able to maintain an accuracy of 83\% on EfficientNet-B5 even as the global batch size increases.

Additionally, we provide a comparison of communication costs and throughput as the global batch size and number of TPU cores scales up in Table~\ref{tab:all_reduce_percent}. We can see that as we increase the number of cores and thus the global batch size, the throughput scales up linearly. In other words, step time remains approximately the same at scale, which may be promising if we wish to scale up even further.


\begin{table}[htb!]
  \caption{\label{tab:all_reduce_percent}Comparison of communication costs and throughput on EfficientNet-B2 and B5 as the global batch size scales up.}
  \label{communication}
  \centering
  \resizebox{0.8\textwidth}{!}{ %
  \begin{tabular}{c cccc}
    \toprule
    \textbf{Model} &
    \makecell{\textbf{\#TPU-v3}\\\textbf{cores}} &
    \makecell{\textbf{Global}\\\textbf{batch size}} &
    \makecell{\textbf{Throughput}\\\textbf{(images/ms)}} &
    \makecell{\textbf{Percent of time}\\\textbf{spent on All-Reduce}}\\
    \midrule
    \multirow{4}{*}{EfficientNet-B2} & 128 & 4096 & 57.57 &2.1    \\
    & 256 & 8192 & 113.73 &2.6   \\
    & 512 & 16384 & 227.13 &2.5    \\
    & 1024 & 32768 & 451.35 &2.81   \\
    \midrule
    \multirow{4}{*}{EfficientNet-B5} & 128 & 4096 & 9.76 & 0.89    \\
    & 256 & 8192 & 19.48 & 1.24 \\
    & 512 & 16384 & 38.55 & 1.24 \\
    & 1024 & 32768 & 77.44 & 1.03 \\
    \bottomrule
  \end{tabular}
  } %
\end{table}

\begin{table}[htb!]
  \caption{\label{tab:peak_acc}Benchmark of EfficientNet-B2 and B5 peak accuracies}
  \label{accuracy}
  \centering
  \setlength\tabcolsep{3pt}
  \resizebox{0.99\textwidth}{!}{ %
  \begin{tabular}{l cccccc c}
    \toprule
    \thead{\textbf{Model}} &
    \thead{\textbf{\#TPU-v3}\\\textbf{cores}} &
    \thead{\textbf{Global}\\\textbf{batch}\\\textbf{size}} &
    \thead{\textbf{Optimizer}} &
    \thead{\textbf{LR}\\\textbf{per 256}\\\textbf{examples}} &
    \thead{\textbf{LR decay}} &
    \thead{\textbf{LR warmup}} &
    \thead{\textbf{Peak}\\\textbf{top-1}\\\textbf{acc.}}\\
    \midrule
    \multirow{5}{*}{EfficientNet-B2}
    & 128 & 4096 & RMSProp & 0.016 & Exponential over 2.4 epochs & 5 epochs & 0.801   \\
    & 256 & 8192 & RMSProp & 0.016 & Exponential over 2.4 epochs & 5 epochs & 0.800   \\
    & 512 & 16384 & RMSProp & 0.016 & Exponential over 2.4 epochs & 5 epochs & 0.799   \\
    & 512 & 16384 & LARS & 0.236 & Polynomial & 50 epochs & 0.795   \\
    & 1024 & 32768 & LARS & 0.118 & Polynomial & 50 epochs & 0.797   \\
    \midrule
    \multirow{6}{*}{EfficientNet-B5}
    & 128 & 4096 & RMSProp & 0.016 & Exponential over 2.4 epochs & 5 epochs & 0.835   \\
    & 256 & 8192 & RMSProp & 0.016 & Exponential over 2.4 epochs & 5 epochs & 0.834   \\
    & 512 & 16384 & RMSProp & 0.016 & Exponential over 2.4 epochs & 5 epochs & 0.834   \\
    & 512 & 16384 & LARS & 0.236 & Polynomial & 50 epochs & 0.833   \\
    & 1024 & 32768 & LARS & 0.118 & Polynomial & 50 epochs & 0.832   \\
    & 1024 & 65536 & LARS & 0.081 & Polynomial & 43 epochs & 0.830   \\
    \bottomrule
  \end{tabular}
  } %
\end{table}
\section{Future Work}
We hope to conduct a deeper study on other large batch optimizers for EfficientNet, such as the SM3 optimizer~\citep{sm3}, in an effort to further improve accuracy at large batch sizes. In addition, model parallelism is a future area of exploration which would supplement the current data parallelism to allow training on large numbers of chips without standard global batch sizes.

\newpage
\bibliographystyle{plainnat}
\bibliography{refs}


\end{document}